\documentclass[letterpaper, 10 pt, conference]{ieeeconf}
\IEEEoverridecommandlockouts
\usepackage{xargs} 
\usepackage{graphicx}
\usepackage{color,soul}
\usepackage{pgfgantt}
\usepackage{pdflscape}
\usepackage{lipsum}
\usepackage{pstricks}
\usepackage{import}
\usepackage{color}
\usepackage{xcolor}
\definecolor{midgrey}{RGB}{100, 100, 100}
\definecolor{darkgreen}{RGB}{0, 100, 50}
\usepackage{layouts}
\usepackage{booktabs}
\usepackage{multirow}
\usepackage{siunitx}
\usepackage{amsmath,amssymb}
\DeclareMathOperator{\E}{\mathbb{E}}
\DeclareMathOperator*{\argmax}{argmax} 

\usepackage{algorithm}
\usepackage[noend]{algpseudocode}
\usepackage{algpseudocode}
\usepackage{stackengine}

\usepackage{pgf}
\usepackage[acronym]{glossaries}
\newacronym{RL}{RL}{reinforcement rearning}

\makeatletter
\def\input@path{{./figures/}}
\makeatother

\makeatletter
\newcommand\fs@spaceruled{\def\@fs@cfont{\bfseries}\let\@fs@capt\floatc@ruled
	\def\@fs@pre{\vspace{0.5\baselineskip}\hrule height.8pt depth0pt \kern2pt}%
	\def\@fs@post{\kern2pt\hrule\relax}%
	\def\@fs@mid{\kern2pt\hrule\kern2pt}%
	\let\@fs@iftopcapt\iftrue}
\makeatother

\makeatletter
\edef\@figdir{_pgfcache}

\catcode`\#=11
\catcode`\|=6
\newcommand{\@basicpgfpreamble}[1]{%
    \unexpanded{%
        \documentclass{standalone}^^J
        \usepackage{pgf}^^J
        \let\oldpgfimage\pgfimage^^J
        \renewcommand{\pgfimage}[2][]{\oldpgfimage[#1]{|1/#2}}^^J
    }%
}
\catcode`\#=6
\catcode`\|=12

\let\@pgfpreamble\@basicpgfpreamble

\newcommand{\setpgfpreamble}[1]{%
    \renewcommand{\@pgfpreamble}[1]{\@basicpgfpreamble{##1}\unexpanded{#1}}
}

\newcounter{@pgfcounter}
\newwrite\@pgfout
\newread\@pgfin

\newcommand{\importpgf}[3][]{%
    \IfFileExists{#2/#3}{}{\errmessage{importpgf: File #2/#3 not found}}%
    \edef\@figfile{\jobname-\the@pgfcounter}%
    \providecommand{\@writetempfile}{}%
    \renewcommand{\@writetempfile}[1]%
    {%
        \immediate\openout\@pgfout=##1%
        \immediate\write\@pgfout{\@pgfpreamble{#2}}%
        \immediate\write\@pgfout{\string\begin{document}}%
        \immediate\openin\@pgfin=#2/#3%
        \begingroup\endlinechar=-1%
            \loop\unless\ifeof\@pgfin%
                \readline\@pgfin to \@fileline%
                \ifx\@fileline\@empty\else%
                    \immediate\write\@pgfout{\@fileline}%
                \fi%
            \repeat%
        \endgroup%
        \immediate\closein\@pgfin%
        \immediate\write\@pgfout{\string\end{document}}%
        \immediate\closeout\@pgfout%
    }%
    \def\@compile%
    {%
        \immediate\write18{pdflatex -interaction=batchmode -output-directory="\@figdir" \@figdir/\@figfile.tex}%
    }%
    \IfFileExists{\@figdir/\@figfile.pdf}%
    {%
        \@writetempfile{\@figdir/tmp.tex}%
        \edef\@hashold{\pdfmdfivesum file {\@figdir/\@figfile.tex}}%
        \edef\@hashnew{\pdfmdfivesum file {\@figdir/tmp.tex}}%
        \ifnum\pdfstrcmp{\@hashold}{\@hashnew}=0%
            \relax%
        \else%
            \@writetempfile{\@figdir/\@figfile.tex}%
            \@compile%
        \fi%
    }%
    {%
        \@writetempfile{\@figdir/\@figfile.tex}%
        \@compile%
    }%
    \IfFileExists{\@figdir/\@figfile.pdf}%
    {\includegraphics[#1]{\@figdir/\@figfile.pdf}}%
    {\errmessage{Error during compilation of figure #2/#3}}%
    \stepcounter{@pgfcounter}%
}
\makeatother

\usepackage[colorinlistoftodos,prependcaption,textsize=tiny]{todonotes}
\newcommandx{\unsure}[2][1=]{\todo[linecolor=red,backgroundcolor=red!25,bordercolor=red,#1]{#2}}
\newcommandx{\change}[2][1=]{\todo[linecolor=blue,backgroundcolor=blue!25,bordercolor=blue,#1]{#2}}
\newcommandx{\info}[2][1=]{\todo[linecolor=OliveGreen,backgroundcolor=OliveGreen!25,bordercolor=OliveGreen,#1]{#2}}
\newcommandx{\improvement}[2][1=]{\todo[linecolor=Plum,backgroundcolor=Plum!25,bordercolor=Plum,#1]{#2}}
\newcommandx{\thiswillnotshow}[2][1=]{\todo[disable,#1]{#2}}
\newcommand{\rework}[1]{%
	\todo[color=yellow,inline, caption={}]{\normalsize Rework: {#1}}%
}

\graphicspath{{./pictures/}{./figures/}}

\newcommand{\figurepath}{./figures}

\newcommand{\tablepath}{./tables}

\newcommand{\Astar}{$\text{A}^* $ }

\usepackage{scalerel,stackengine,amsmath}
\newcommand\equalhat{\mathrel{\stackon[1.5pt]{=}{\stretchto{%
				\scalerel*[\widthof{=}]{\wedge}{\rule{1ex}{3ex}}}{0.5ex}}}}

\usepackage[style=ieee, bibencoding=utf8,	sorting=none, maxcitenames=2, mincitenames=1, minbibnames=2, maxbibnames=2, backend=biber, citestyle=numeric-comp, isbn=false,doi=false, url=false, natbib=true]{biblatex}

\AtEveryBibitem{%
	\ifentrytype{book}{
		\clearfield{url}%
		\clearfield{urldate}%
		\clearfield{review}%
		\clearfield{series}
	}{}
	\ifentrytype{collection}{
		\clearfield{url}%
		\clearfield{urldate}%
		\clearfield{review}%
		\clearfield{series}
	}{}
	\ifentrytype{incollection}{
		\clearfield{url}%
		\clearfield{urldate}%
		\clearfield{review}%
		\clearfield{series}
	}{}
		\ifentrytype{inproceedings}{
		\clearfield{url}%
		\clearfield{urldate}%
		\clearfield{review}%
		\clearfield{series}
		\clearfield{publisher}
	}{}
}

\DeclareNameAlias{sortname}{last-first}
\DeclareNameAlias{default}{last-first}

 \usepackage[caption=false,font=footnotesize]{subfig}

 \usepackage{dblfloatfix}

\newcommand\copyrighttext{%
	\scriptsize \textcolor{blue}{\textcopyright 2018 IEEE. Personal use of this material is permitted.  Permission from IEEE must be obtained for all other uses, in any current or future media, including reprinting/republishing this material for advertising or promotional purposes, creating new collective works, for resale or redistribution to servers or lists, or reuse of any copyrighted component of this work in other works}}
\newcommand\copyrightnotice{%
	\begin{tikzpicture}[remember picture,overlay]
	\node[anchor=north,yshift=-7.5pt] at (current page.north) {\fbox{\parbox{\dimexpr\textwidth-\fboxsep-\fboxrule\relax}{\copyrighttext}}};
	\end{tikzpicture}%
}

\begin{document}
%
\title{\begin{center} \LARGE \bf Experience-Based Heuristic Search: Robust Motion Planning\\ with Deep Q-Learning      \end{center}}

%
%
%

%
\author{Julian Bernhard$^{1}$, Robert Gieselmann$^{1}$, Klemens Esterle$^{1}$ and Alois Knoll$^{2}$%
	\thanks{$^{1}$Julian Bernhard, Robert Gieselmann and Klemens Esterle are with fortiss GmbH, An-Institut Technische Universit\"{a}t M\"{u}nchen, Munich, Germany}%
	\thanks{$^{2}$Alois Knoll is with Chair of Robotics, Artificial Intelligence and Real-time Systems, Technische Universit\"{a}t M\"{u}nchen, Munich, Germany}%
}
%


\maketitle
\copyrightnotice
\thispagestyle{empty}
\pagestyle{empty}

\begin{abstract} Interaction-aware planning for autonomous driving requires an exploration of a combinatorial solution space when using conventional search- or optimization-based motion planners.
With Deep Reinforcement Learning, optimal driving strategies for such problems can be derived also for higher-dimensional problems. However, these methods guarantee optimality of the resulting policy only in a statistical sense, which impedes their usage in safety critical systems, such as autonomous vehicles.  
Thus, we propose the Experience-Based-Heuristic-Search algorithm, which overcomes the statistical failure rate of a Deep-reinforcement-learning-based planner and still benefits computationally from the pre-learned optimal policy. Specifically, we show how experiences in the form of a Deep Q-Network can be integrated as heuristic into a heuristic search algorithm. We benchmark our algorithm in the field of path planning in semi-structured valet parking scenarios. There, we analyze the accuracy of such estimates and demonstrate the computational advantages and robustness of our method.
Our method may encourage further investigation of the applicability of reinforcement-learning-based planning in the field of self-driving vehicles.
\end{abstract}

\section{Introduction}

Motion planners for self-driving vehicles frequently adhere to optimization- or search-based paradigms. At each new planning run, these methods reexamine the solution space to find an optimal motion. For higher-dimensional planning scenarios, this is computationally demanding. 
For instance, at the strategic level, such approaches commonly evaluate only a subset of potential maneuvers and their interaction with the traffic scene, restricting their usage to scenarios with a reduced number of participants and a limited time horizon \cite{hubmann_decision_2017, kessler_multi_2017}. In path planning scenarios in unstructured environments, heuristic search algorithms, such as the Hybrid \Astar algorithm, \cite{dolgov_path_2010} fully reexplore the configuration space on every replanning task. 

In contrast, humans rely on their past experiences to evaluate the safety and suitability of a maneuver. This allows them to handle complex planning problems with ease. Inspired by this, with Reinforcement Learning (RL), an optimal policy is derived by exploiting all past environmental interactions. The ongoing success in applying \acrshort{RL} using neural networks to high-dimensional problems \cite{mnih_human-level_2015,van_hasselt_deep_2016} motivated its use for deriving driving policies for intersection crossing \cite{isele_navigating_2017} or highway maneuvering \cite{li_reinforcement_2015}. However, approximate \acrshort{RL} methods guarantee optimality of the learned policy merely in a \textsl{statistical} sense, impeding their usage in safety critical systems such as autonomous vehicles.

\begin{figure}[t]
	\def\svgwidth{\columnwidth}
	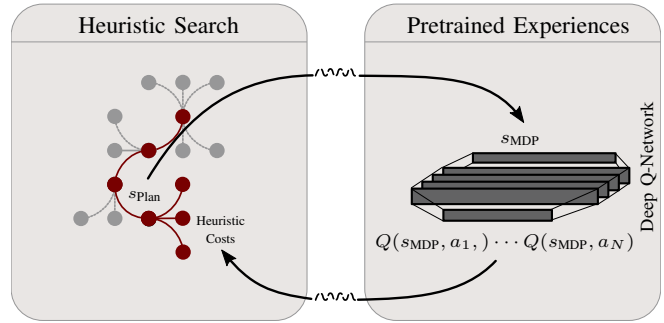
	\caption{The experience-based-heuristic-search algorithm relies on a pretrained Deep Q-Network to guide an incremental search. During node expansion, a single forward pass through the DQN, followed by a postprocessing step, yields the heuristic costs for all expanded child nodes. Compared to baseline approaches, we benefit computationally from the optimal policy encoded within the network, even when the planning state $s_\text{plan}$ and training state $s_\text{MDP}$ are defined differently. }  \label{fig:top_graphic}
	\vspace{-2mm}
\end{figure}

This motivates our work: The Experience-Based Heuristic Search (EBHS) algorithm integrates experiences in the form of pretrained Q-values into a heuristic search as depicted in Figure~\ref{fig:top_graphic}. We demonstrate that our algorithm benefits computationally from the pretrained experiences. Further, it overcomes the statistical failure rate of a pure reinforcement-learning-based planner due to the added search process. 

Specifically, we apply Double Deep Q-Networks \cite{van_hasselt_deep_2016} and learning from demonstration \cite{hester_learning_2017} to learn the state-action values \(Q(s,a)\) for two application types in the field of path planning. The learned Q-functions are integrated into a Hybrid \Astar planner to replace the commonly used heuristic functions.

The main contributions of this paper are:
\begin{itemize}
	\item An adaptation of an heuristic search algorithm to use learned experiences in the form of a Q-function as heuristic estimate. 
	\item The evaluation of variants of Deep-Q-learning algorithms and their parameters to study the accuracy of the derived heuristic estimate.
	\item A demonstration of the computational advantages when using our experience-based planner in semi-structured valet parking scenarios.
	\item A demonstration of the reliability of such an approach compared to pure reinforcement learning based planning.
	
\end{itemize}

The structure of this paper is as follows: First, we present previous work related to our field. Then, we introduce the EBHS algorithm. Next, we present the results of experience learning and, finally, an application to different planning scenarios and a statistical analysis of the robustness of our method.

\section{Related Work} \label{sec:related_work}

The heuristic function plays an important role in all informed search algorithms. Previous work already combined search-based methods with learned heuristic functions, obtained either by supervised or reinforcement learning. 
A combination of Monte Carlo Tree Search (MCTS) with a learned policy and value network led to a mayor breakthrough in artificial intelligence by beating the best human players in the game of Go \cite{silver_mastering_2016}. \citet{paxton_combining_2017} adapt this approach to discrete task planning for autonomous driving. However, as \textsl{continuous} state spaces remain challenging for the MCTS algorithm, their approach impedes a generation of continuous, dynamic behavior.

In the field of heuristic learning, \citet{li_ann:_2016} trained a neural network with supervised learning to estimate a correction factor for a standard heuristic. Yet, their approach cannot replace the actual heuristic function. \mbox{\citet{pareekutty_rrt-hx:_2016}} use value iteration to iteratively create a quality grid map during planning, which guides the node expansion of a RRT planner. However, their approach uses a discretized state space and does not allow pretraining of the heuristic. 

Using imitation learning, \citet{bhardwaj_learning_2017} first acquire an optimal policy for a distribution of potential planning scenarios. This policy is used to guide a best-first search when planning for a specific scenario within the distribution. 
Similar to our approach, they encode the optimal policy with a Q-function. However, as they directly use the policy, instead of calculating a heuristic from the Q-values, their algorithm ignores the optimality of the solution. 

We benchmark our algorithm in the field of path planning in unstructured environments.
A common approach in this field is the Hybrid \Astar algorithm extending the standard \Astar algorithm towards a continuous state representation. It uses the maximum of two different heuristic functions~\cite{dolgov_path_2010}: A holonomic version considering obstacles and a non-holonomic version considering the kinematic constraints. We observed that this heuristic leads to long planning times in certain planning scenarios, since the two sub-heuristics may guide towards contradicting states.

To reduce planning time, the orientation-aware space exploration guided heuristic search algorithm creates a unified heuristic function \cite{chen_motion_2016}. In a pre-planning step, it performs a circle-based state exploration, leading to a decrease in planning time compared to the conventional Hybrid \Astar implementation.
Other ways of heuristic definition are higher cost regions dependent on the amount of required additional gear shifts \cite{liu_boundary_2017} or are based on a separation of the configuration space into visible and non-visible regions \cite{choi_efficient_2012}. The above methods are suitable to decrease planning time in more complex, maze-like environments. In contrast, we investigate, if exploiting an already learned maneuver might be more beneficial to reduce planning time in standard parking maneuvers.
In semi-structured environments with lanes given, planning should consider the road geometry. Up-to now, no analytical heuristic exists which estimates the non-holonomic path onto a curved lane. Instead, with a look-ahead parameter, a configuration on the curve is fixed, forming a planning problem with a single goal configuration~\mbox{\cite{fassbender_motion_2016, chen_motion_2016}}. This parameter, however, does not generalize well to different situations. 

Compared to existing work, we show how a learned Q-function can be used as the \textsl{only} heuristic in an \mbox{\Astar-algorithm} to search for an \textsl{optimal} solution in a \textsl{continuous} state space. We learn a non-holonomic heuristic for semi-structured environments, disregarding obstacles, and a unifying heuristic for standard parking scenarios considering both vehicle constraints and obstacles. Further, we show that a combination of learning and search-based methods benefits from the optimality of the learned policy and the increase in robustness due to the additional search. This may pave the way to practical applications of machine learning algorithms for motion planning algorithms of autonomous vehicles.

\section{Problem Definition}
We want to find the sequence of actions leading from an environment start state \(s_{start}^\text{Plan}\) to one of several possible environment goal states \(\mathcal{S}_g =\{s_{goal,l}^\text{Plan}\}, l \in \{1,\dots,H\}\). The actual end state $s_\text{goal}^{*}$ fulfills an optimality criterion, e.g. giving the path with minimum length. 

The \Astar algorithm finds the minimum cost solution by building a search tree  rooted at \(s_{start}^\text{Plan}\). By applying the set of possible actions \(\mathcal{A} = \left\{ a_i \right\}, i \in \{1,\dots,N\}\) from the current best state, new child states are expanded and the tree is iteratively grown until a goal configuration is reached. In each expansion step, the state with the lowest total cost \(f(s^\text{Plan}) = g(s^\text{Plan}) + h(s^\text{Plan})\) is selected with \(g(s^\text{Plan})\) being the cost from the start state \(s_{start}^\text{Plan}\) to the current state \(s^\text{Plan}\) and \(h(s^\text{Plan})\) naming the cost-to-go metric or heuristic function from the current state \(s^\text{Plan}\) to the set of goal states \(\mathcal{S}_g\). A closed list contains already expanded nodes. The search process is over either when the open list is empty or the number of maximum iterations is reached.

To ensure fast convergence of the search, the following conditions should hold for a heuristic function \(h(s^\text{Plan})\):
\begin{itemize}
	\item Admissibility \(h(\cdot) \leq h_{opt}(\cdot)\): \(h(\cdot)\) should never overestimate the true cost-to-go \(h_{opt}(\cdot)\).
	
	\item Optimality $ h(\cdot)  \approx h_{opt}(\cdot) $: If $h(\cdot)$ is close to the true cost-to-go value, this fastens goal expansion and reduces processing time. 

\end{itemize}
 
We propose a learning-based mechanism to meet these requirements. 

\section{Experience-Based Heuristic Search}
We derive how a state-action value \(Q(s^\text{MDP}, a)\) yields a heuristic function \(h(s^\text{Plan})\) in the EBHS algorithm. In the following derivation, we set \(s\equalhat s^\text{MDP}\) for better readability. 

\subsection{Q-Learning}
Reinforcement learning seeks an optimal policy for the problem of sequential decision making formulated as Markov Decision Process (MDP). One distinguishes between value-based and policy-gradient methods. Q-learning belongs to the category of model-free, value-based reinforcement learning methods \cite{mnih_human-level_2015}. It learns the state-action value function
\begin{equation} \label{eq:true_qvalue}
Q^\pi(s,a) = \E_{\pi}\left[\sum_{t=0}^\infty \gamma^tr_t| s_0=s,a_0=a\right],
\end{equation}
representing the expected return, taking action \(a\) in state \(s\) and from thereon following policy \(\pi\). The discount factor \(\gamma\) defines how future rewards $r_t$ contribute to the current state-action value.
The Bellman equation 
\begin{equation} \label{eq:bellman}
Q^*(s,a) = \E_{s^\prime}\left[r(s,a,s^\prime) + \gamma \max_{a^\prime} Q^*(s^\prime,a^\prime)|s,a\right]
\end{equation}
defines the fix point of the optimal action-value function from which the optimal policy \(a^*= \pi^*(s) = \argmax_a Q^*(s,a)\) is derived.

\subsection{Q-function Integration}
The MDP and planning state definitions may differ. A problem-dependent transformation $s^\text{MDP}=t(s^\text{Plan})$ links the two state definitions. 
\subsubsection{Definition of the Rewards}
\begin{figure}[b]
	\centering
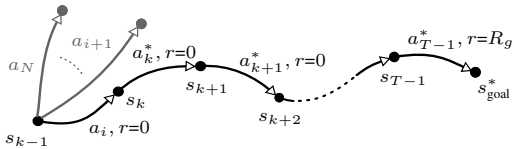
\caption{Visualization of the state transitions when following the optimal policy $a^*_k =\pi^*(s_{k})$ to the goal. We require a sparse reward setting to enable a straightforward calculation of the heuristic function from the Q-function.} \label{fig:state_transitions}
\end{figure}

The reward definition of the MDP shall simplify the heuristic calculation from the Q-function. As we will derive in the following, this requires
\begin{equation*}
 r(s,a,s^\prime) = \begin{cases}
 		R_{\text{g}}, & \text{if } s^\prime \in t(\mathcal{S}_g) \\
 		0, & \text{otherwise,}
 \end{cases}
\end{equation*} meaning the only non-zero reward is given for a transition onto a goal state. Figure \ref{fig:state_transitions} visualizes this \textsl{sparse} reward setting.

\subsubsection{Preserving the Greedy Policy}
For the following reasoning, we assume that \(g(s_k) = 0\), meaning during node expansion, the search algorithm ignores passed way-costs. In this \textsl{theoretical} setting, the order of expanded nodes of the EBHS algorithm shall resemble the state sequence of the optimal policy \(\pi^*\). 

We achieve this, by establishing a inversely proportional relationship between the heuristic value \(h(s_k^\text{Plan})\) and the Q-function of the parent state \(Q(s_{k-1},a_i)\) with the action \(a_i\) leading from \(s_{k-1}\) to \(s_k\). Figure \ref{fig:state_transitions} shows the corresponding state transitions. 
For instance, if \(a_i\) is optimal in state \(s_{k-1}\), meaning \({a_i = \text{argmax}_a Q(s_{k-1},a)} \), the heuristic \(h(s_k^\text{Plan})\) shall take the lowest value among all nodes expanded from \(s_{k-1}\).

\subsubsection{Calculation of the Heuristic}
By combining the reward setting in aspect 1) with the Q-function definition in equation \ref{eq:true_qvalue}, we get
\begin{equation} \label{eq:heuristic_calc}
\begin{split}
&Q^*(s_{k-1},a_i)  = 0 + \gamma Q^*(s_k,a^*_k) \\
&= 0 + \gamma \big[0 + \gamma  Q^*(s_{k+1},a^*_{k+1}) \big]\\
&= 0 + \gamma \big[0 +  \gamma \big[ \dots + \gamma \big[0 + \gamma \underbrace{ Q^*(s_{T-1},a^*_{T-1})}_{= R_\text{g}} \big]\big]\big]\\ 
&= \gamma^L \cdot R_{\text{g}}
\end{split}
\end{equation}

 where \(L\) is the number of steps from state \(s_k\) to the goal. Solving equation \(\ref{eq:heuristic_calc}\) for \(L\) yields the heuristic estimate 
\begin{equation}
 h(s_k^\text{Plan}) = L \cdot c_a = \log_{\gamma} \frac{Q^*(s_{k-1}^\text{MDP},a_i)}{R_{\text{g}} } \cdot c_{a}\,.
\end{equation}
We require a unique cost value \(c_{a}\)  for all motion segments. 

\subsection{Deep Q-Networks for Heuristic Learning}

\floatstyle{spaceruled}
\restylefloat{algorithm}
\begin{algorithm} [t]
	
	\caption{ExpandNodeEBHS(\(s_{k-1}^\text{Plan}\)) } \label{alg:node_expansion}

	\label{alg:ebhs_algorithm}
	\begin{algorithmic}[1]
		\State \(\mathcal{S}_\text{children} \leftarrow \varnothing\)
		\State $s_{k-1}^\text{MDP}=t(s_{k-1}^\text{Plan})$
		\State	$\text{qvalues}_{1,\ldots,N} = \text{ForwardEvaluationDQN}(s_{k-1}^\text{MDP})$
		\For{$a_i \in \mathcal{A}, i \in 1,\dots,N$ }
		\State $ s_{\text{child},i}^\text{Plan} = \text{SimulateMotionSegment}(s_{k-1}^\text{Plan},a_i)$
			\If{!colliding($s_{\text{child},i}^\text{Plan}$)}
			\State  $f(s_\text{child,i}^\text{Plan}) = g(s_\text{child,i}^\text{Plan}) +  \log_{\gamma} \frac{\text{qvalues}_{i}}{R_{\text{g}} } \cdot c_{a}   $
		\State $ \mathcal{S}_\text{children} \leftarrow s_\text{child,i}^\text{Plan}$
			\EndIf
		\EndFor
		\State Output: $\mathcal{S}_\text{children}$
	\end{algorithmic}	
\end{algorithm}

To enable planning in a continuous state space, we must represent \(Q(s,a)\) with a function approximator, such as a neural network.  \citet{mnih_human-level_2015} successfully applied Q-learning to problems with higher dimensional continuous state spaces. To overcome divergence issues when using neural networks for Q-function representation, they introduced the concepts of a target network and an experience replay buffer. 

For integration of an \textsl{approximated} Q-function \(\widetilde{Q}(s,a)\) as heuristic function, it is important to minimize its difference to the true Q-value \(Q^*(s,a)\) given in equation \ref{eq:true_qvalue}. Further, the training algorithm must be capable of dealing with sparse reward settings. Therefore, we evaluate the following algorithmic adaptations of DQN:
\begin{itemize}
	\item \textsl{Double Deep Q-learning} (DDQN) \cite{van_hasselt_deep_2016} aims to reduce the upward bias inherent to approximated Q-values. This bias arises due to the maximum operation within the Bellman update.  
	\item \textsl{Prioritized experience replay} \cite{schaul_prioritized_2016} gives increased priority to experiences with high temporal-difference (TD) error, improving convergence, especially in sparse reward settings. 
	\item We apply \textsl{\(n\)-step Deep Q-learning} proposed by \citet{mnih_asynchronous_2016}, but in a synchronous version. It reduces the upward bias of the Q-value estimate and fastens the propagation of rewards to previously visited states. However, convergence is impeded due to higher variances of the TD-error estimates.  
	\item Learning from  demonstrations becomes beneficial when dealing with high-dimensional state spaces and sparse rewards. Thus, we apply \textsl{Deep Q-learning from Demonstrations} (DQfD) \cite{hester_learning_2017} which allows pretraining from an expert policy while still preserving the Bellman property.
\end{itemize} 
Further algorithmic details are found in the respective publications.

We employ the standard neural network architecture for DQNs, outputting a vector of Q-values for all actions. We benefit computationally from this architecture, as we require only a single forward evaluation of the DQN in the node expansion step, to retrieve the heuristic costs for all children. Algorithm \ref{alg:node_expansion} describes the node expansion process of the EBHS algorithm.

\section{Experiment}
We benchmark the EBHS algorithm in two applications from the field of path planning: 
\begin{itemize}
	\item \textbf{Non-holonomic heuristic learning (NHL):}
	As discussed in section \ref{sec:related_work}, for semi-structured scenarios, no suitable non-holonomic heuristic exists for planning onto a continuously-curved road segment. Thus, we learn a non-holonomic heuristic estimating the optimal path onto a quadratic Bezier curve. The slope at a specific point on the curve defines the desired vehicle orientation at this point. Obstacles are considered by the EBHS algorithm and not during experience learning.
	\item \textbf{Learning of a unified heuristic (UHL):}
	We learn a unified heuristic for a standard parking scenario. The learned policy considers both non-holonomic constraints and obstacles. The scenario consists of two rows of four parking spaces placed opposite each other. The start configuration is arbitrarily oriented and placed between these rows. The goal is positioned in one of the eight parking spaces, and oriented forwards or backwards.
\end{itemize}

\subsection{Experience Learning}

We show how the experiences in form of a Deep Q-function were acquired for the two applications and discuss findings of the training processes.

\subsubsection{MDP Definition}
The vehicle kinematics were described by a single track model with discretized steering angle $\kappa$ and a constant speed $v$ for forward and backward motions. 
We used the same motion primitives \({a_i = \left\{\kappa_i,  v\right\}}\) for the RL agent, as used later on by the Hybrid $\text{A}^*$ algorithm for graph expansion. 

We evaluated two types of representations of the vehicle configuration: a standard form with a normalized orientation \({ c_{s}=(x_s,y_s,\theta/2\pi)}\), and a trigonometric version \({ c_{t}=(x_s,y_s,sin(\theta),cos(\theta))}\). The latter avoids a value jump after a full turn, which proved to be more beneficial in the UHL setting. The coordinate values \(x,y\) were normalized with respect to the workspace boundaries.

Positive rewards were given when the vehicle reached a tolerance region around the goal, negative rewards for collisions with the workspace boundary, or in case of the UHL setting, when colliding with an occupied parking lot. An episode was over either after colliding or when reaching the maximum number of allowed actions.
Table \ref{ref:rl_problems} in the appendix provides the detailed MDP definitions.

\begin{figure}
	
	\centering
	\subfloat[NHL]{	\includegraphics{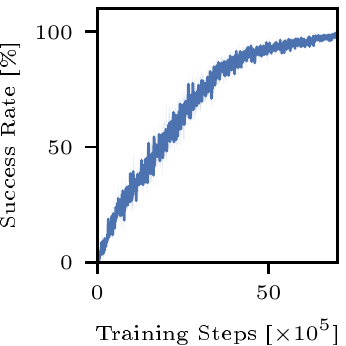}} 
	\subfloat[UHL]{	\includegraphics{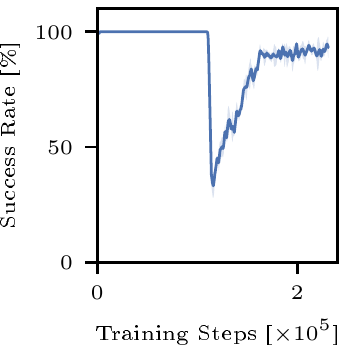} \label{fig:final_reward_parking}} 
	
	\caption{The learning curves for the best performing hyperparameter sets for the NHL and UHL setting averaged over three training runs. NHL used prioritized DDQN with one-step return. Due to the sparse reward setting a long training time was required. In the UHL setting, we employed a DQfD algorithm. It initializes the policy from demonstrations, explaining the high success rate at the beginning.}  \label{fig:final_reward}
	\vspace{-2mm}
\end{figure}

\subsubsection{Deep Reinforcement Learning} \label{sec:algorithms}
For the NHL application, we applied prioritized DDQN \cite{schaul_prioritized_2016} and experimented with different \(n\)-step returns \cite{mnih_asynchronous_2016}. For the UHL application, we found that an initialization of the policy from expert demonstrations is beneficial to ease exploration of the higher dimensional state space. Therefore, we employed prioritized DQfD \cite{hester_learning_2017} with Hybrid \Astar expert demonstrations. For both cases, we used \(\epsilon\)-greedy exploration.

\subsubsection{Network Architectures}
The Q-function was approximated by a fully connected network with \(u\) hidden ReLU layers. A linear layer for NHL and a tanh layer for UHL with size $N$ outputted a state-action value for each of the motion primitives. The input layer had the dimensions of the MDP state space.

\subsubsection{Training and Test Data}
The initial states of the MDPs were sampled from fixed training data sets at the beginning of each episode. To obtain a data set for the NHL setting, we fixed the first two Bezier curve supporting points in the left half of the workspace. Then, we sampled 100 Bezier curves by moving the third point on a half circle in the right half of the workspace. Each of the 100 Bezier curves was combined with 1000 randomly sampled vehicle start configurations resulting in a randomized but fixed training set with \num{e5} MDP states.
For the UHL setting, we separated the  training data into goal and start vehicle configurations and combined these sets randomly during training. The goal set consisted of one forward and one backward vehicle configuration for each of the eight parking spaces. To obtain the start configurations, we defined a grid in the configuration space with spacings \({\Delta x=\SI{0.3}{\metre}, \Delta y=\SI{0.3}{\metre},\Delta\theta=\SI{30}{\degree}}\) and sorted out all colliding configurations. Combining this two configuration sets, gave a training set with roughly \num{6e4} MDP states. Our equal sized test set for UHL consists of all intermediate start configurations in between the training configurations.

\subsubsection{Results}
A random search over the most relevant hyperparameters was performed to improve the success rate of the learned policies. The success rate describes how often the goal configuration is reached on average over the last episodes. Figure \ref{fig:final_reward} shows the success rates over the course of training for the best-performing parameters. 
Table \ref{ref:rl_problems} in the appendix summarizes the most relevant parameters used in the final evaluation.

\subsection{Studying the Accuracy of Heuristic Estimates}
In a next step, we investigated the effect of certain hyperparameters on the accuracy of the heuristic estimate for NHL. To simplify notation in the following, we use \(h(s_k)=h(s_k^\text{Plan})\).

\begin{figure}[t] 
	\centering
	\vspace{2mm}
	\subfloat[Single Step Difference]{\includegraphics{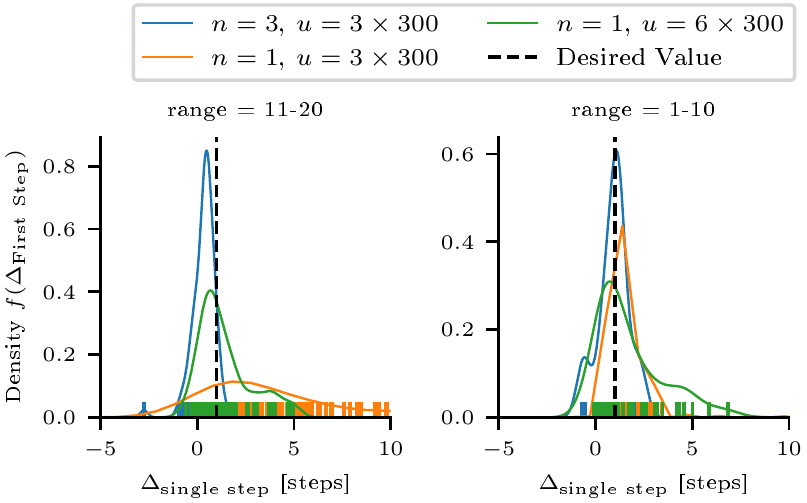} \label{fig:single_step}} \\
	\subfloat[Total Difference]{\includegraphics{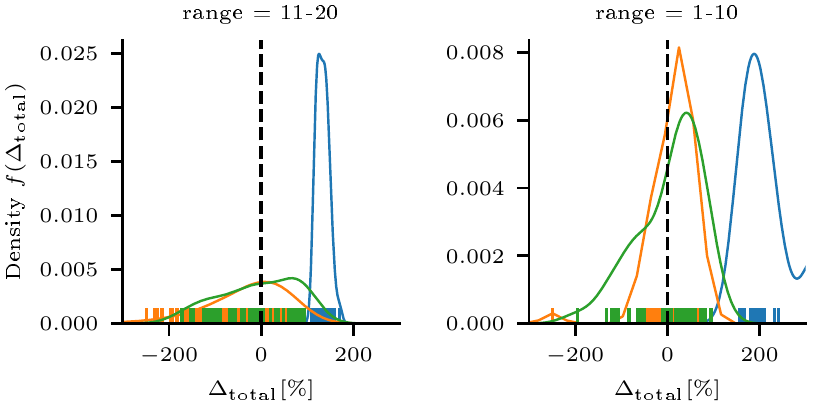}}\\
	\subfloat[Training Results]{ \scriptsize  \begin{tabular}{lll}
\toprule
           Parameters & Success Rate [\%] & Normalized TAE [\%] \\
\midrule
 $n=3$, $u=3\times300$ &              97.2 &               249.1 \\
 $n=1$, $u=3\times300$ &             99.0 &                 6.8 \\
 $n=1$, $u=6\times300$ &              97.8 &                16.4 \\
\bottomrule
\end{tabular}
  \label{tab:heuristic_results}}
	\caption{Density estimates of the heuristic accuracy metrics for different hyperparameters ($u$: number of layers, $n$: length of $n$-step return) and remaining steps to the goal (range) in the NHL setting. Below, we give the corresponding final success rates and TAEs.} \label{fig:heuristic_eval}
	\vspace{-2mm}
\end{figure}

According to \citet{anschel_averaged-dqn:_2017}, the difference between the optimal Q-function \(Q^*(s,a)\), defined by the Bellman equation, and the Q-function \(\widetilde{Q}(s,a)\) approximated by a neural network can be decomposed into:
\begin{itemize}
 \item The \textsl{target approximation error} (TAE): During training, we minimize the temporal differences between subsequent state-action pairs. The TAE is the remaining minimization error after training. It arises due to inexact optimization of the loss functions, finite capacity of the neural network and insufficient generalization to unseen state-action pairs.
 \item The  \textsl{overestimation error} (OE): Noise during environment interaction leads to overestimations of the Q-values due to the maximum operation in the Bellman equation (Equation \ref{eq:bellman}). Double Deep Q-Networks reduce this effect.  
But, as discussed in  \cite{anschel_averaged-dqn:_2017}, a growth in TAE variance, a higher number of actions or increasing the discount factor heightens this type of error. The variance of the TAE is reduced with larger \(n\)-step return, as we bootstrap further in the future to estimate the temporal difference. 
 \item The \textsl{optimality difference} is the error between standard tabular Q-learning and the optimal Q-function \(Q^*(s,a)\). It is negligible in our evaluation. 
\end{itemize}
To make different parameter settings comparable in our evaluation, we define a normalized TAE error as \[\text{TAE}_\text{Norm} = \frac{\text{TAE}}{R_\text{g} \cdot n}.\] We divide by the goal reward, and, as the temporal difference error sums up with the \(n\)-step return, also by $n$. 
For different hyperparameters, Table \ref{tab:heuristic_results} depicts the normalized TAE after training, averaged over a mini-batch of training samples, and the corresponding final success rates.
Though, the size of the action set and the discount factor influence the TAE, a change of these parameters greatly affected the success rate of the learned policy. Thus, these parameters were left out in this evaluation. We observe that the success rates resemble each other for the different parameter settings, but the TAE varies greatly. 

To study the effects of the remaining TAE on the accuracy of the heuristic estimates, we defined two evaluation metrics. The single step difference
\begin{equation*}
  \Delta_\text{single step} = \widetilde{h}(s_k) - \widetilde{h}(s_{k+1})
 \end{equation*} 
expresses how the learned heuristic estimate \(\widetilde{h}(\cdot)\) changes from a parent state \(s_k\) to a child state \(s_{k+1}\). 
The total relative difference 
\begin{equation*}
 \Delta_\text{total} = \frac{h_\text{opt}(s_k) - \widetilde{h}(s_k)} {h_\text{opt}(s_k)}
\end{equation*}
compares the true cost-to-go \(h_\text{opt}(\cdot)\) to the learned heuristic estimate \(\widetilde{h}(\cdot)\). To increase convergence speed of the heuristic search algorithm, $\Delta_\text{single step}$ should be slightly lower than the cost of a motion segment \(c_a\). $\Delta_\text{total}$ should be close to zero.

We estimate probability densities of this metrics using \num{e4}~training samples, in which the learned policy successfully reached the goal. For each of these samples, we calculated the true cost-to-go \(h_\text{opt}(\cdot)\) by multiplying the motion cost with the true number of required steps, obtained by following the learned policy to the goal. The child state for calculation of $\Delta_\text{single step}$ was obtained by applying the greedy action from the parent state. Figure \ref{fig:heuristic_eval} compares the density estimates of these metrics for the NHL setting for different hyperparameters.

The goal of our evaluation was to clarify why certain settings worked better in the final evaluation than others. For the three analyzed parameter settings, we summarize our main observations as follows:
\begin{itemize}
	\item \(n=3\), \(u=3\times300\): We obtain accurate peaks at the desired value for $\Delta_\text{single step}$, and, as expected, the distribution has small variance. However, $\Delta_\text{total}$ shows a large offset. We assume that this is due to the high remaining TAE.
	\item \(n=1\), \(u=3\times300\): We observe average performance for $\Delta_\text{single step}$, but best performance for $\Delta_\text{total}$, however, with an overall increase in variance.
	\item \(n=1\), \(u=6\times300\): We expected the lowest TAE, however, unstable training amplified the TAE. The densities peak near the desired values, but the distributions are non-Gaussian. We assume overfitting of the Q-function, which lead to larger errors at non-frequently visited states.
\end{itemize}
Considering \textsl{both} evaluation metrics one-step DQN with small network capacity performed best. Thus, we selected its learned Q-function for NHL in the final evaluation. 

We proposed two metrics which served as guidance for selecting suitable hyperparameters for experience learning. Yet, a profound study in the future should refine our metric definition and evaluate the influence of the observed variances on the performance of the EBHS algorithm.   

\subsection{Final Evaluation of EBHS}\label{baseline}

In a final evaluation of the EBHS algorithm, we want to approach the following questions:
\begin{itemize}
	\item How well can the EBHS algorithm benefit computationally from the pretrained experiences in comparison to baseline approaches? 
	\item Can the EBHS algorithm generalize to scenarios not covered by the MDP state definition?
	\item Can the statistical failure rate of a pure reinforcement-learning-based approach b\texttt{}e overcome with the EBHS algorithm?
\end{itemize}

\subsubsection{Implementation Aspects}
We use the C++ implementation of the Hybrid A* algorithm presented in \cite{chen_motion_2016}. We interface with a tensorflow-based implementation in Python to estimate the heuristic costs for one expansion step and return them to the planner. 

As baseline heuristic, we employ for the UHL application only the Reeds-Shepp heuristic. The additional \Astar heuristic worsened the performance of the baseline Hybrid \Astar in our experiment. We disable direct Reeds-Shepp goal expansion~\cite{dolgov_path_2010}, as both EBHS and the Hybrid $\text{A}^*$ would benefit from it.
For the NHL application, no analytical heuristic exists for planning onto a quadratic Bezier curve. Thus, we approximate it as follows: We sample the goal Bezier curve at equidistant points and calculate a Reed Shepp path to each of them. Then, we take the minimum-length path as heuristic cost. 

We performed the final evaluation on an Intel Core i7~@~3.3~GHz and 16 GB Ram with \textsl{disabled} graphic card support to ensure same processing conditions for all of the approaches.

\subsubsection{Scenario Evaluation}
We applied the EBHS algorithm and the baseline approaches to two scenarios for the NHL and UHL setting. For the NHL application, we selected a pullout maneuver and a parallel marking maneuver onto a curved road. Note that the obstacles in the NHL scenarios are not part of the MDP state space. The UHL setting was evaluated on a reverse parking scenario from the training data set, and a scenario where we added an obstacle \textsl{not} considered by the MDP state definition.

\begin{figure}

	\centering
	\subfloat[NHL: Parallel Pullout]{	\includegraphics{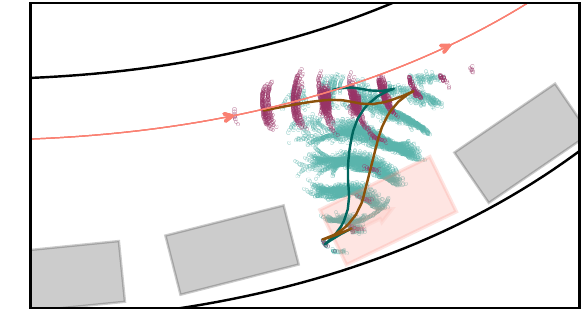}}\\ \vspace{-2mm}%
	\subfloat[NHL: Backwards Pullout]{\includegraphics{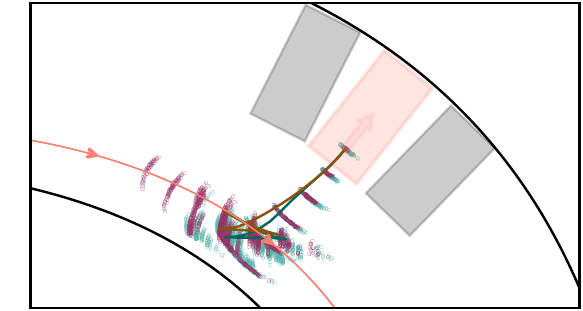}}\\ \vspace{-2mm}
	\subfloat[UHL: Additional Obstacle]{\includegraphics{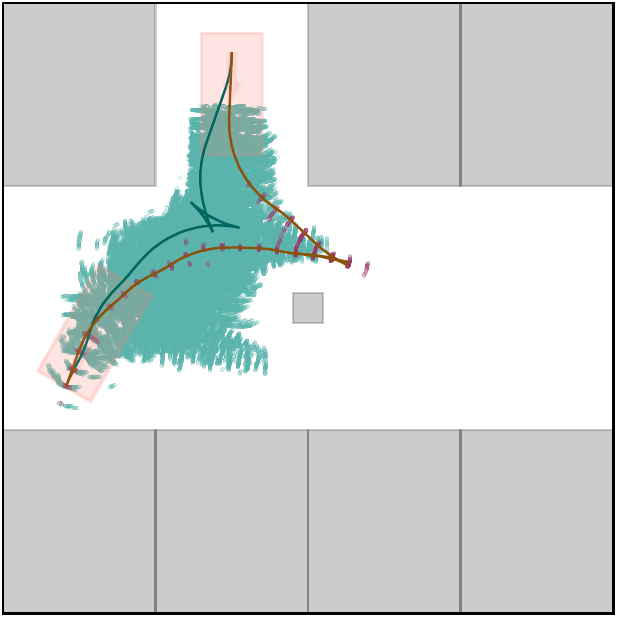} }\\ \vspace{-2mm}
	\subfloat[UHL: Reverse]{\includegraphics{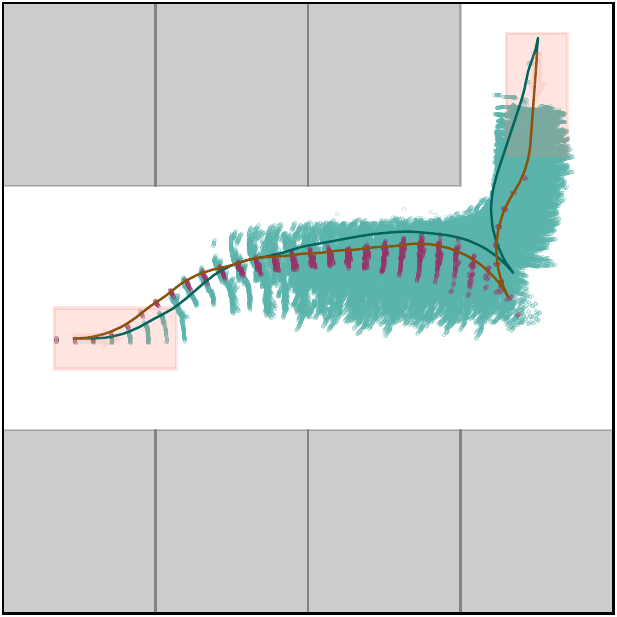}}\\
	{\includegraphics{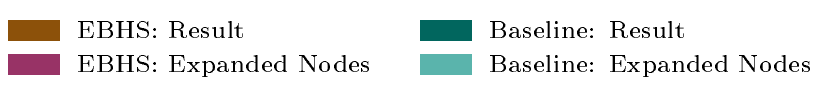} } \vspace{-2mm}
	\caption{Visualization of planned paths and expanded nodes for EBHS and baseline planners for the NHL and UHL application. EBHS led to a lower number of expanded nodes and faster planning time.}
\label{fig:scenario_results}
\end{figure}

For this scenarios, Figure \ref{fig:scenario_results} shows the resulting paths and expanded nodes. Tables \ref{tab:bezier} and \ref{tab:parking} provide numerical results. For all depicted scenarios, EBHS required a significantly lower number of planning iterations and a lower planning time, though, one iteration in EBHS is computationally more costly due to the forward evaluation of the DQN and the Python interfacing.
When adding an obstacle not included in the MDP state, the EBHS algorithm still benefited from the pretrained experiences indicating the generalization capabilities of our approach.
However, EBHS generated a longer path in this case. For the future, we plan to investigate how to improve optimality of the solution in generalization scenarios.

\subsubsection{Statistical Evaluation}

\begin{figure}[t]
	\centering
	\vspace{2mm}
	\includegraphics{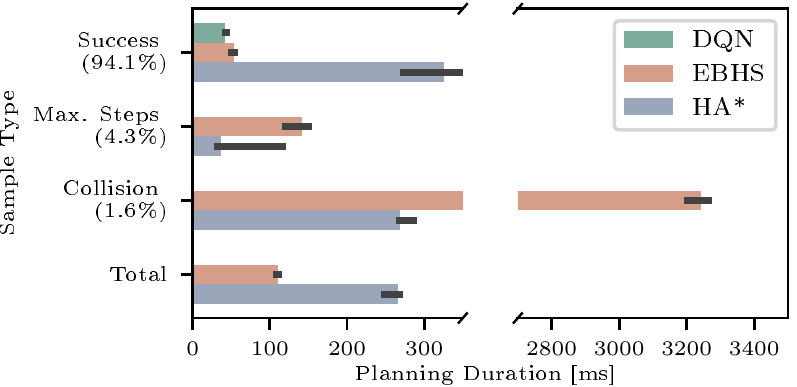} \vspace{-2mm}
	\caption{Median planning durations with confidence bounds for the UHL scenario, estimated from 1000 test samples for each sample category, for a pure DQN-based planner, the EBHS algorithm and the Hybrid \Astar baseline. The EBHS algorithm outperformed the baseline over the total test set and even succeeded for samples, in which the DQN-based planner collided or reached the maximum number of allowed steps. } 	\label{tab:statisical_eval}
	\vspace{-3mm}
\end{figure}

Figure \ref{fig:final_reward_parking} depicts a success rate of 90\% at the end of training in the UHL setting. Hence, learning of a suitable policy failed for 10\% of the training data. For these fail samples, the learned policy either exceeded the maximum number of steps, due to the learned policy getting trapped in local minima, or lead to a collision in case of difficult starting positions near the workspace boundary. This stochastic failure behavior occurs due to the learned policy optimizing the \textsl{expectation} of the return over \textsl{all} states visited during training. 

To see, if EBHS can overcome the stochastic failure rate, we compared the planning durations of the EBHS algorithm, a pure DQN-based planner and the baseline Hybrid $\text{A}^*$ separately for fail and success \textsl{test} samples. The test data showed an equal failure rates as the training data. Figure \ref{tab:statisical_eval} shows the resulting medians with confidence bounds for UHL estimated using 1000 samples for each category.

For the success samples, EBHS and DQN clearly outperformed the Hybrid $\text{A}^*$. Thus, we conclude that the node expansion process in the EBHS algorithm mainly follows the learned policy and spends only slight computational overhead with unnecessary node expansions. In contrast to DQN,  EBHS always found a solution for the failure samples, but with higher planning duration than the Hybrid A*. The total median duration over the whole test set outperforms the Hybrid $\text{A}^*$ by $60\%$. 

Our evaluation showed that the EBHS algorithm successfully exploits learned experiences to speed up the convergence of the search process. The search process itself ensures robustness against the statistical failure rate of a pure DQN-based planner. 

\begin{table}
	\vspace{2mm}
	\caption{Planning Results for non-holonomic heuristic learning (NHL)} \label{tab:bezier}
	\centering
	\begin{tabular}{lllll}
\toprule
Scenario & \multicolumn{2}{l}{Parallel Pullout} & \multicolumn{2}{l}{Backwards Pullout} \\
Planner &             EBHS & Baseline &              EBHS & Baseline \\
\midrule
Planning Time [s] &              5.2 &      8.1 &               3.2 &     13.8 \\
Expanded Nodes    &             2096 &     3467 &              2080 &     9115 \\
Iterations        &              503 &     1630 &               227 &     1954 \\
Path Length       &              9.0 &      9.0 &              12.0 &      9.0 \\
\bottomrule
\end{tabular}

\end{table}
\begin{table}[t]
	\caption{Planning Results for unified heuristic learning (UHL)} \label{tab:parking}
	\centering
	\begin{tabular}{lllll}
\toprule
Scenario & \multicolumn{2}{l}{Added Obstacle} & \multicolumn{2}{l}{Reverse} \\
Planner &           EBHS & Baseline &    EBHS & Baseline \\
\midrule
Planning Time [s] &            0.2 &      3.9 &     0.5 &      4.5 \\
Expanded Nodes    &            778 &   175230 &    2567 &   205862 \\
Iterations        &            137 &   132682 &     457 &   131602 \\
Path Length       &           26.6 &     16.9 &    23.9 &     24.3 \\
\bottomrule
\end{tabular}

	\vspace{-2mm}
\end{table}

\section{Conclusion and Future Work}
We presented the EBHS algorithm, which uses experiences in the form of a Deep Q-Network as heuristic function in a heuristic search, and proposed two metrics to assess the accuracy of learned heuristic estimates for different hyperparameter settings. We empirically proved that, with an additional search, we overcome the statistical failure rate of Deep-reinforcement-learning-based planning, but still benefit computationally from a pre-learned optimal policy.

\citet{silver_mastering_2016} demonstrated the advantages of combining reinforcement learning with search-based algorithms for planning in discrete state spaces. The EBHS algorithm represents now a step forward in applying this principle to continuous state spaces. Yet, a better understanding of the DQN overestimation errors and the accuracy of the learned heuristics could further increase the benefits of our method. 

In the future, we plan to further investigate the generalization capabilities of our method, and apply it to strategic planning tasks in dynamic environments.

\begin{table*}[t]
	\vspace{2mm}
	\caption{Summary of most relevant hyperparameters for learning the experiences with Deep Q-Networks.} \label{ref:rl_problems}
	\renewcommand{\arraystretch}{0.9}
\centering	\begin{tabular}{@{}p{3.7cm}p{6cm}p{6cm}@{}}
		\toprule[2pt]
		& \textbf{Non-holonomic Heuristic Learning (NHL)} & \textbf{Unified Heuristic Learning (UHL)} \\ \midrule[1.5pt]
		\textbf{MDP Definition} \\
		State Space & $s_\text{MDP} = \left(c_{\text{start},s}, P_1,P_2,P_3\right) \in \mathbb{R}^9$ with Bezier curve supporting points $P_i=(x_i,y_i)$ &$s_\text{MDP} = \left( c_{\text{start},t}, c_{\text{goal},t}, o_1, o_2,\ldots, o_8\right) \in \mathbb{R}^{16}$ with one-hot encoding $ o_i \in \left\{ 0, 1\right\} $ of parking spaces\\
		Work Space [m] & $ 0 \leq x \leq 30$, $ 0 \leq y \leq 30$ &  $0 \leq x \leq 20$, $ 0 \leq y \leq 20$ \\
		Action Space  & $\kappa = \pm 30^\circ, \pm 20^\circ, \pm 10^\circ, 0^\circ $, $v = \pm 5.0 \frac{m}{s}$  &  $\kappa = \pm 17.2^\circ, \pm 8.6^\circ, 0^\circ $, $v = \pm 3.0 \frac{m}{s}$\\
		Reward & goal: $+1000$, collision: $-1000$ & goal: $+1$, collision: $-1$ \\
		Time Step  & $0.2\, s$ & $0.2\, s$   \\
		Discount Factor & 0.95 & 0.95 \\
		Transition Model & deterministic: Single Track Vehicle & deterministic: Single Track Vehicle \\
		\midrule[0.5pt]
		\textbf{Deep Reinforcement Learning} \\
		Algorithm & Prioritized DDQN  & Prioritized DQfD with Hybrid $\text{A}^*$ demonstrations \\
		Length of \(n\)-step return &1 &5 \\
		Hidden ReLU Layers x Units& 3x300 & 5x300 \\
		Output Layer Type & Linear & Tanh \\
		\bottomrule[2pt]
	\end{tabular}

\end{table*}


%

\appendices


\footnotesize
\printbibliography

%


\end{document}